\documentclass[journal]{IEEEtran}

\usepackage{cite}
\usepackage{amsmath,amssymb}
\usepackage{graphicx}
\usepackage[T1]{fontenc}
\usepackage[utf8]{inputenc}
\usepackage{siunitx}
\DeclareMathOperator{\AMI}{AMI}
\DeclareMathOperator{\MI}{MI}
\DeclareMathOperator{\ent}{H}

\title{Comprehensive Deployment-Oriented Assessment for Cross-Environment Generalization in Deep Learning-Based mmWave Radar Sensing}

\author{Tomoya~Tanaka, Tomonori~Ikeda, and Ryo~Yonemoto%
\thanks{Tomoya Tanaka is with the School of Electrical and Computer Engineering, Georgia Institute of Technology, Atlanta, GA 30332 USA (e-mail: tomoya.tanaka@gatech.edu).}%
\thanks{Tomoya Tanaka, Tomonori Ikeda, and Ryo Yonemoto are with SoftBank Corp., Tokyo, Japan (e-mail: \{tomoya.tanaka, tomonori.ikeda, ryo.yonemoto\}@softbank.jp).}%
}

\begin{document}
\maketitle

\begin{abstract}
This study presents the first comprehensive evaluation of spatial generalization techniques, which are essential for the practical deployment of deep learning-based radio-frequency (RF) sensing. Focusing on people counting in indoor environments using frequency-modulated continuous-wave (FMCW) multiple-input multiple-output (MIMO) radar, we systematically investigate a broad set of approaches, including amplitude-based statistical preprocessing (sigmoid weighting and threshold zeroing), frequency-domain filtering, autoencoder-based background suppression, data augmentation strategies, and transfer learning. Experimental results collected across two environments with different layouts demonstrate that sigmoid-based amplitude weighting consistently achieves superior cross-environment performance, yielding 50.1\% and 55.2\% reductions in root-mean-square error (RMSE) and mean absolute error (MAE), respectively, compared with baseline methods. Data augmentation provides additional though modest benefits, with improvements up to 8.8\% in MAE. By contrast, transfer learning proves indispensable for large spatial shifts, achieving 82.1\% and 91.3\% reductions in RMSE and MAE, respectively, with 540 target-domain samples. Taken together, these findings establish a highly practical direction for developing radar sensing systems capable of maintaining robust accuracy under spatial variations by integrating deep learning models with amplitude-based preprocessing and efficient transfer learning.
\end{abstract}

\begin{IEEEkeywords}
amplitude weighting, cross-environment generalization, data augmentation, 
FMCW MIMO radar, mmWave radar sensing, people counting, transfer learning
\end{IEEEkeywords}


\section{Introduction}
In real-world environments, there is a growing demand for sensing technologies that can detect human presence and activities in a non-contact and robust manner. Radio Frequency (RF) sensing has emerged as a promising solution, offering advantages over vision-based approaches such as privacy preservation, resilience to lighting and weather conditions, and the ability to penetrate opaque objects. These characteristics make RF sensing suitable for stable operation in a wide range of indoor and outdoor scenarios~\cite{Gu2016_Sensors_Review,Li2013_TMTT_Healthcare,RecentReview2023_mmWave}.

Among various RF sensing methods, Multiple-Input Multiple-Output (MIMO) radar systems are particularly attractive, as they enable simultaneous estimation of the distance and angle of arrival (AoA) of reflected signals~\cite{Endo2023_MIMO_PositionEstimation}. This capability allows for precise localization and accurate people counting. In recent years, research efforts have focused on integrating such sensor data with deep learning models—such as Convolutional Neural Networks (CNNs) and Long Short-Term Memory (LSTM) networks—to achieve higher-level semantic understanding~\cite{Will2019_JSEN_24GHz,Ren2023_JIOT_GroupedCount,Lin2023_Sensors_Multitask,Zhang2019_LSENS_HAR,mmPose2020_JSEN}.

However, while deep learning-based methods demonstrate high accuracy under controlled conditions, their generalization to unseen spatial environments remains limited due to their susceptibility to overfitting~\cite{OpenRadar2021_RadarConf}. Unlike domains such as computer vision or natural language processing, RF sensing inherently faces challenges in large-scale data collection. Sensor deployment and human-assisted data acquisition are required for each environment, making it difficult to adopt web-scale data-driven approaches~\cite{Seyfioglu2017_IGRSL_Init,GAN_Augmentation2023,PhysAugment2023}. Additionally, physical factors such as wall materials, room geometry, and furniture layout significantly affect electromagnetic propagation, leading to pronounced domain shifts across different settings~\cite{OpenRadar2021_RadarConf,MultipathMitigation2023_TIM,MultipathMapping2024_JIOT,Permittivity2021_APS}. Ensuring stable performance under such constraints is therefore crucial for practical RF sensing applications~\cite{CrossEnvHAR2022_JSEN,Mauro2023_ApplIntell_FSL}.

From the perspective of prior work, existing studies can be broadly divided into two categories: (1) those that train and evaluate solely within the same environment, without assessing the impact of environmental changes~\cite{Zhang2019_LSENS_HAR,mmPose2020_JSEN}; and (2) those that evaluate in environments different from the training domain, thereby explicitly examining the influence of spatial variation. The present work falls into the second category. Within (2), some studies have evaluated the effect of transfer learning~\cite{Mauro2023_ApplIntell_FSL,MetaLearning2022_ICASSP}, while others have examined the effect of data augmentation~\cite{GAN_Augmentation2023,PhysAugment2023}, as well as preprocessing strategies such as autoencoder-based background suppression before model training~\cite{Autoencoder_ICASSP2020,Ghifary2016_Autoencoder_Domain}. However, to the best of our knowledge, no previous study has provided a unified evaluation of preprocessing, data augmentation, and transfer learning, all of which affect spatial variation, under the same experimental conditions. Such a comprehensive evaluation is indispensable for clarifying the mechanisms required for highly practical deep learning-based RF sensing, and in this respect, this study establishes a significant milestone toward achieving deployable, spatially robust radar sensing systems.

To address these challenges, this study proposes a comprehensive evaluation framework to improve the spatial generalization of people counting models based on MIMO radar. Specifically, we systematically evaluate the effectiveness of three complementary approaches under identical experimental conditions: (1) an advanced preprocessing method that suppresses background noise while enhancing human-related reflections, (2) a systematic data augmentation scheme to improve learning diversity and robustness under limited data conditions, and (3) an efficient transfer learning that enables rapid adaptation with minimal labeled samples from the target domain.

\section{Experimental Setup}

This study employs the same radar hardware as in our concurrent research on model comparison. However, the present work focuses on spatial generalization across varied environments, using independently collected data and evaluating preprocessing and transfer learning strategies.

\subsection{Radar Hardware and Signal Processing}

We used a millimeter-wave MIMO radar system based on the FMCW (Frequency Modulated Continuous Wave) principle, which linearly sweeps its carrier frequency and analyzes the frequency difference between transmitted and received signals to estimate both target distance and relative motion. In particular, FMCW radars generate beat signals that are converted into range profiles through Fast Fourier Transform (FFT), followed by further angular resolution processing such as Synthetic Aperture (SA) techniques.

The radar used in our experiments operates at \SI{24.15}{\giga\hertz} and is equipped with two transmit and four receive antennas, allowing 2D reflection intensity maps (range vs. azimuth) to be generated with a resolution of $12 \times 91$. These maps provide spatial information about reflected signal amplitude, from which human presence can be inferred.

The exact radar hardware configuration, processing pipeline, and experimental setup details—such as antenna placement, environmental layouts, and data formatting—are thoroughly documented in our earlier technical report published by IEICE~\cite{ieiceRadarTechRep}.

 \subsection{Environment Configuration}

This study evaluates spatial generalization through a systematic three-stage framework:
\begin{itemize}
    \item \textbf{Stage 1: Intra-layout Validation} -- Basic performance verification within identical configurations
    \item \textbf{Stage 2: Layout-level Adaptation} -- Adaptation to furniture rearrangement within the same physical environment
    \item \textbf{Stage 3: Space-level Adaptation} -- Generalization performance evaluation across different physical environments
\end{itemize}

This hierarchical approach enables isolation of specific generalization challenges and targeted solution development.

To evaluate this three-stage framework, we prepared three distinct environments, as summarized in Table~\ref{tab:env_specs}.

\begin{table}[htp]
\centering
\caption{Environment Specifications Comparison}
\label{tab:env_specs}
\renewcommand{\arraystretch}{1.2}
\begin{tabular}{p{1.8cm} p{1.8cm} p{1.8cm} p{1.8cm}}
\hline
\textbf{Specification} & \textbf{Env A (Stage 1)} & \textbf{Env B (Stage 2)} & \textbf{Env C (Stage 3)} \\
\hline
Room Size & \SI{4.9}{\meter}~$\times$~\SI{6.9}{\meter} & \SI{4.9}{\meter}~$\times$~\SI{6.9}{\meter} & \SI{6.3}{\meter}~$\times$~\SI{6.3}{\meter} \\
\hline
Ceiling Height & \SI{2.7}{\meter} & \SI{2.7}{\meter} & \SI{2.9}{\meter} \\
\hline
Wall Material & RF Absorber & RF Absorber & Plasterboard + Vinyl Paint \\
\hline
Floor Material & Carpet & Carpet & Carpet \\
\hline
Major Furniture & 0--1 items & 6 items & 10 or more items \\
\hline
Radar Position & \((x, y, z)\) & \((x, y, z)\) & \((x', y', z)\) \\
\hline
Acoustic Property & Anechoic & Anechoic & Reverberant \\
\hline
\end{tabular}
\end{table}

Environment~A consists of four layout patterns in a darkroom setting: no furniture, one to four single chairs only, two desks only, and one whiteboard only. 

Environment~B is located in the same darkroom but includes three chairs, two desks, and one whiteboard. It was designed to simulate a scenario in which the spatial layout changes while the physical location remains constant. The radar installation in Environment~B is identical to that in Environment~A, ensuring that only layout changes, not sensor placement, contribute to performance differences. 

The purpose of evaluating Environment~B is to assess how effective preprocessing and data augmentation techniques are in maintaining model performance under layout variation. This evaluation scenario simulates real-world deployment situations where furniture arrangements and spatial configurations may change over time due to operational requirements, renovations, or daily usage patterns, while the physical location and sensor installation remain constant.

In contrast, Environment~C represents a completely different physical setting, with changes in both room structure and furniture/equipment layout compared to the training environment. It was designed to assess the effectiveness of transfer learning in environments with distinct spatial configurations, requiring rapid adaptation for practical model deployment.

\begin{figure}[htp]
\centering
\includegraphics[width=1.0\linewidth]{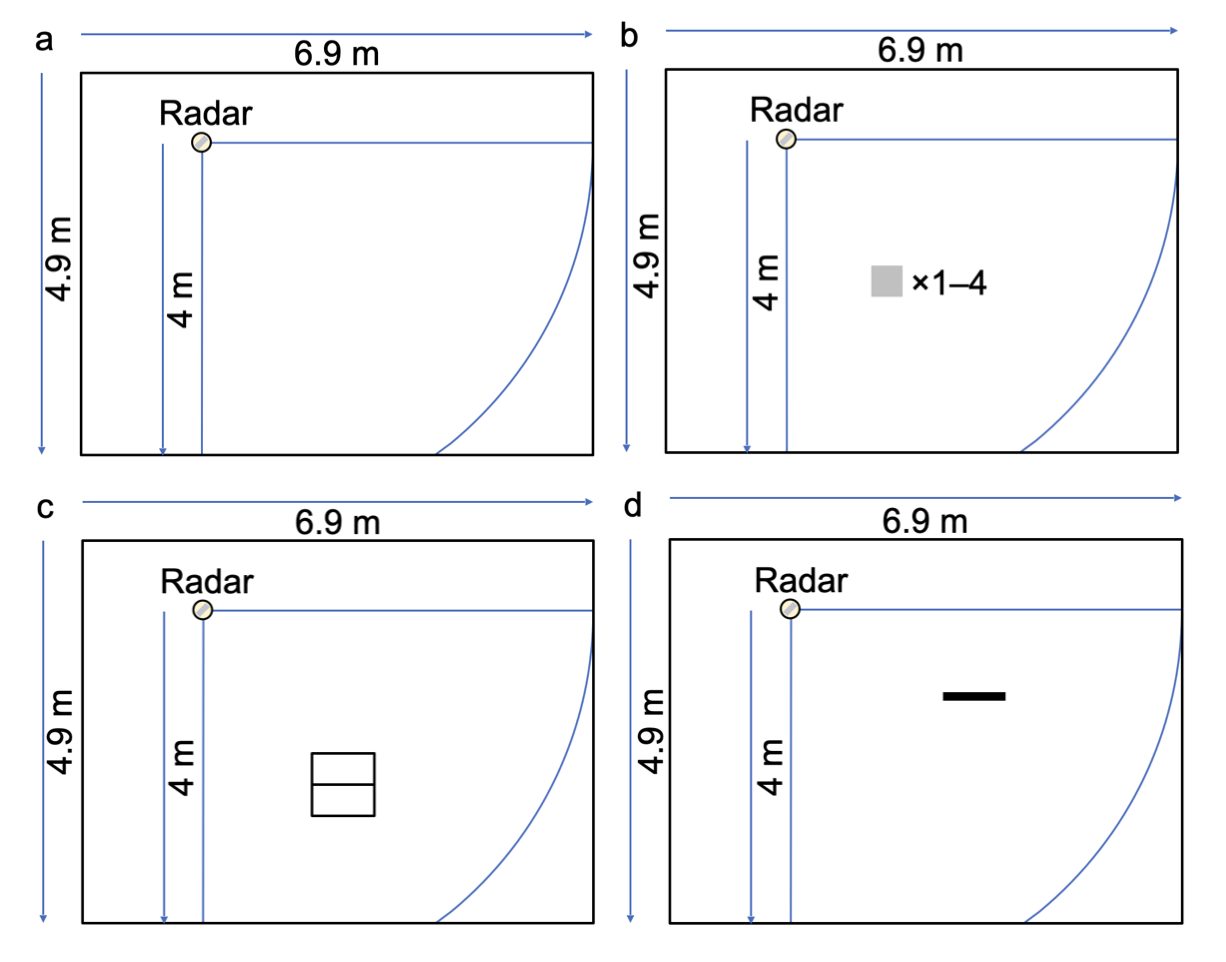}
\caption{ Layout variations for Environment~A: (a) empty, (b) random placement of 1--4 single chairs (gray squares indicate an example chair arrangement and do not necessarily represent the actual positions), (c) two fixed tables (unfilled rectangles indicate table positions), (d) one fixed whiteboard (black narrow rectangle indicates its position). The fan-shaped area from the radar shows its detection range, up to \SI{5}{\meter}.
}
\label{fig:envA}
\end{figure}

\begin{figure}[htp]
\centering
\includegraphics[width=0.6\linewidth]{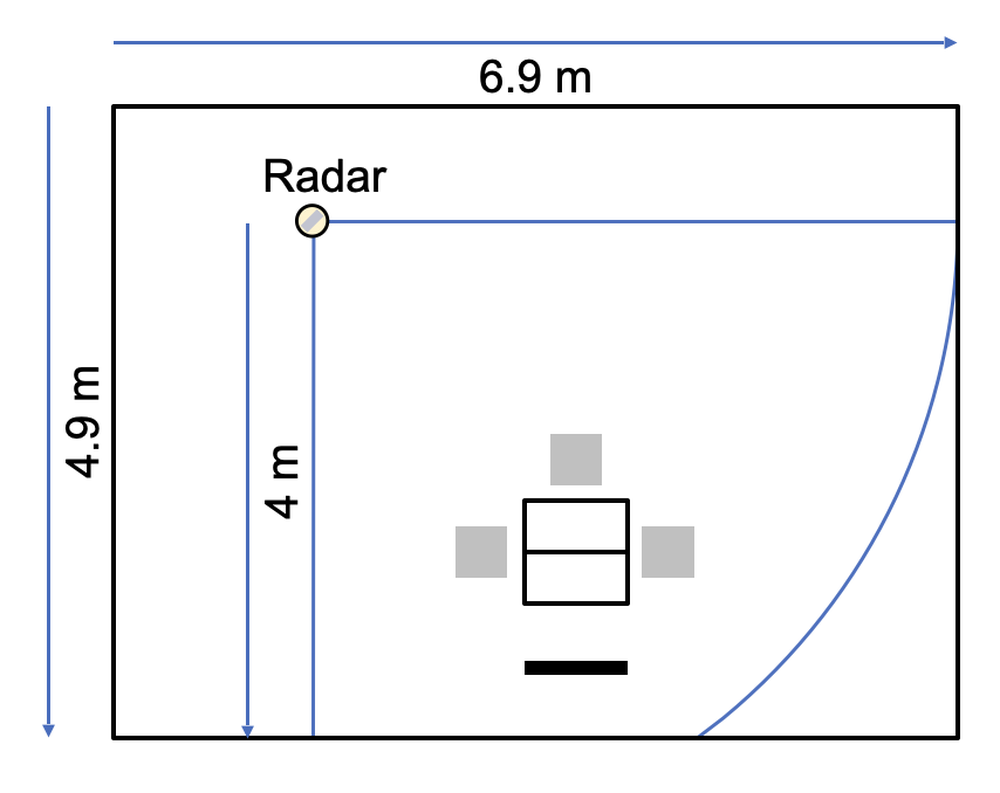}
\caption{Layout of Environment B: same chamber as Environment A, with three chairs, two tables, and one whiteboard. The layout shown in the figure represents the actual positions.}
\label{fig:envB}
\end{figure}

\begin{figure}[htp]
\centering
\includegraphics[width=0.6\linewidth]{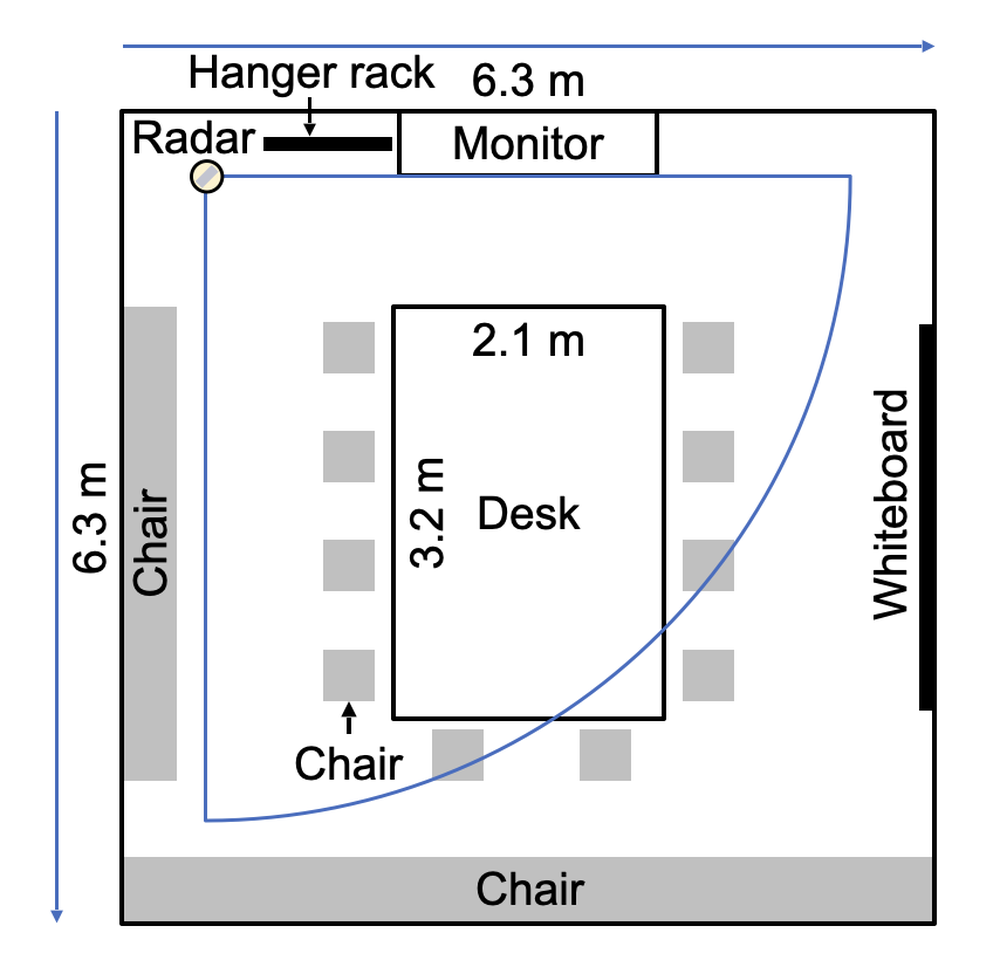}
\caption{Layout of Environment C: a different meeting room with distinct size, height, wall materials, and furniture configuration.}
\label{fig:envC}
\end{figure}

The layouts of Environments A, B, and C are illustrated in Figures~\ref{fig:envA}, \ref{fig:envB}, and \ref{fig:envC}. 

\subsection{Dataset Overview}

Table~\ref{tab:dataset} summarizes the number of samples collected in Environments A, B, and C. Each radar sample is represented as a $12 \times 91 \times 60$ array, corresponding to range bins, angular bins, and temporal frames ($\approx$\SI{7}{\second}). Each pixel denotes signal amplitude. 

\begin{table}[htp]
\caption{Number of samples per class in each environment.}
\label{tab:dataset}
\centering
\begin{tabular}{c c c c}
\hline
\textbf{Class (Number of Persons)} & \textbf{Env A} & \textbf{Env B} & \textbf{Env C} \\
\hline
0 & 1600 + 2200$^\dagger$ & 400 & 250 \\
\hline
1 & 1600 & 400 & 250 \\
\hline
2 & 1600 & 400 & 250 \\
\hline
3 & 1600 & 400 & 250 \\
\hline
\end{tabular}

\footnotesize{$^\dagger$2200 additional samples used for training the autoencoder.}
\end{table}

Three adult male participants (30s–40s) took part in the data collection. Each sample included one of three activity states—standing still, walking irregularly, or a mixture of both—representing typical office and meeting room scenarios. These patterns comprehensively cover static and dynamic human presence.

Environment A served as the primary dataset for model training and performance evaluation, with samples split into training, validation, and test sets. Additional samples from the 0-person class were used exclusively for autoencoder training. Environment B was used for spatial generalization evaluation, and Environment C for transfer learning evaluation. All radar data underwent preprocessing: Outlier Clipping (0.1st–99.9th percentiles) to remove noise, and Min-Max Normalization to [0,1] to account for sensor and environment variability.
 
\section{Generalization Enhancement Methods}

\subsection{Baseline CNN-LSTM Model}

\begin{figure}[htp]
\centering
\includegraphics[width=1.0\linewidth]{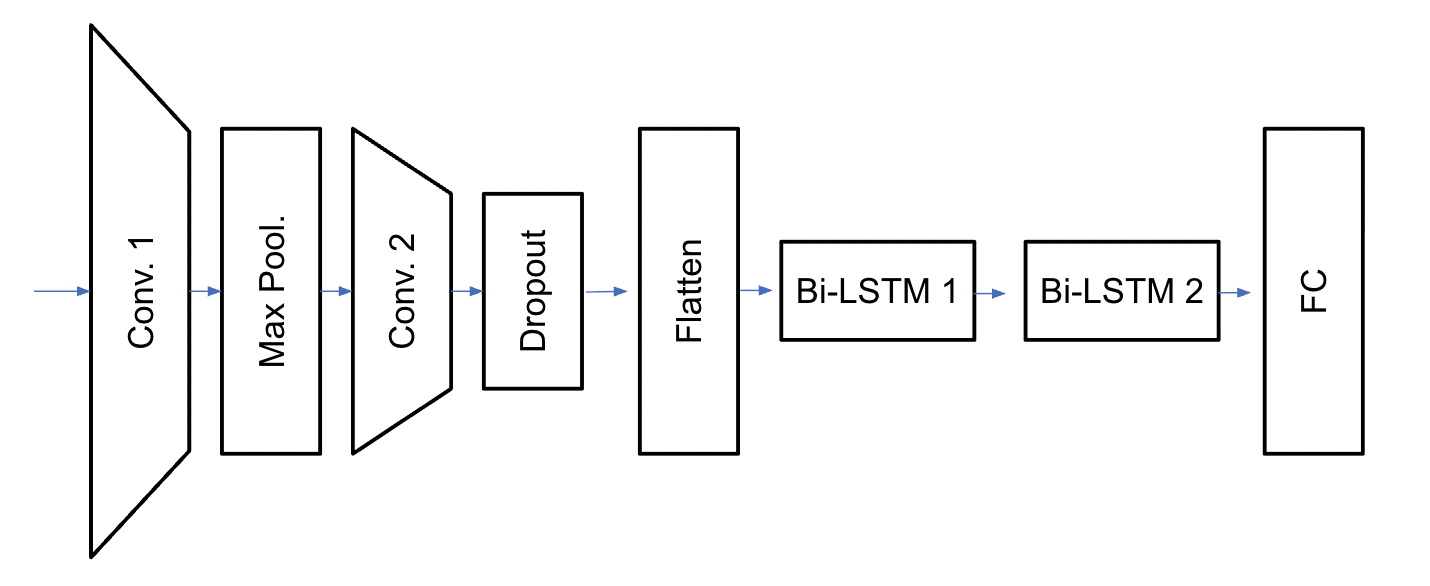}
\caption{Architecture of the baseline CNN-LSTM model. The CNN block extracts spatial features from each radar frame, while the Bi-LSTM block models temporal patterns across the sequence.}
\label{fig:model_architecture}
\end{figure}

As illustrated in Fig.~\ref{fig:model_architecture} and summarized in Table~\ref{tab:model_layers}, the baseline model combines a CNN for spatial feature extraction with a bidirectional LSTM (Bi-LSTM) for temporal modeling. The input is a 60-frame radar sequence ($12 \times 91$), corresponding to about \SI{7}{\second} of acquisition, which provides sufficient temporal context for capturing motion-related amplitude variations and physiological cues.  

The baseline CNN-LSTM model extracts spatial features from each radar frame via two convolutional layers with pooling and dropout, followed by a two-layer Bi-LSTM (128 units) to capture temporal dynamics. A final fully connected layer regresses the number of people (0--3). Training uses MSE loss with the Adam optimizer (learning rate $10^{-3}$) and early stopping (patience 10). This model provides the reference framework for evaluating preprocessing, data augmentation, and transfer learning.

This CNN-LSTM serves as the baseline framework for evaluating the impact of preprocessing, data augmentation, and transfer learning on spatial generalization.

\begin{table}[htp]
\caption{Layer configuration of the baseline CNN-LSTM model.}
\label{tab:model_layers}
\centering
\small
\renewcommand{\arraystretch}{1.2}
\begin{tabular}{c p{3.0cm} p{3.0cm}}
\hline
\textbf{Layer} & \textbf{Description} & \textbf{Output Shape} \\
\hline
Input & 60-frame sequence ($12 \times 91$) & $B \times 60 \times 1 \times 12 \times 91$ \\
\hline
Conv1 & Conv2D ($1 \rightarrow 16$, $3\times3$) + ReLU & $B \times 60 \times 16 \times 12 \times 91$ \\
\hline
MaxPool & MaxPool2D ($2\times2$) & $B \times 60 \times 16 \times 6 \times 45$ \\
\hline
Conv2 & Conv2D ($16 \rightarrow 32$, $3\times3$) + ReLU & $B \times 60 \times 32 \times 6 \times 45$ \\
\hline
Flatten & Flatten for LSTM input & $B \times 60 \times 8640$ \\
\hline
Bi-LSTM & 2-layer, 128 units, bidirectional & $B \times 60 \times 256$ \\
\hline
FC & Fully connected (256 $\rightarrow$ 1) & $B \times 1$ \\
\hline
\end{tabular}
\end{table}

\subsection{Pre-processing}
\subsubsection{Processing Based on the Standard Deviation of Amplitude}
Human reflections show temporal fluctuations from breathing, heartbeat, and micro-movements, whereas static objects remain nearly constant~\cite{KimLing2009_TGRS}. To exploit this difference, we used the temporal standard deviation of amplitude values to separate human-related signals from background reflections.

\begin{figure}[htp]
\centering
\includegraphics[width=0.47\textwidth]{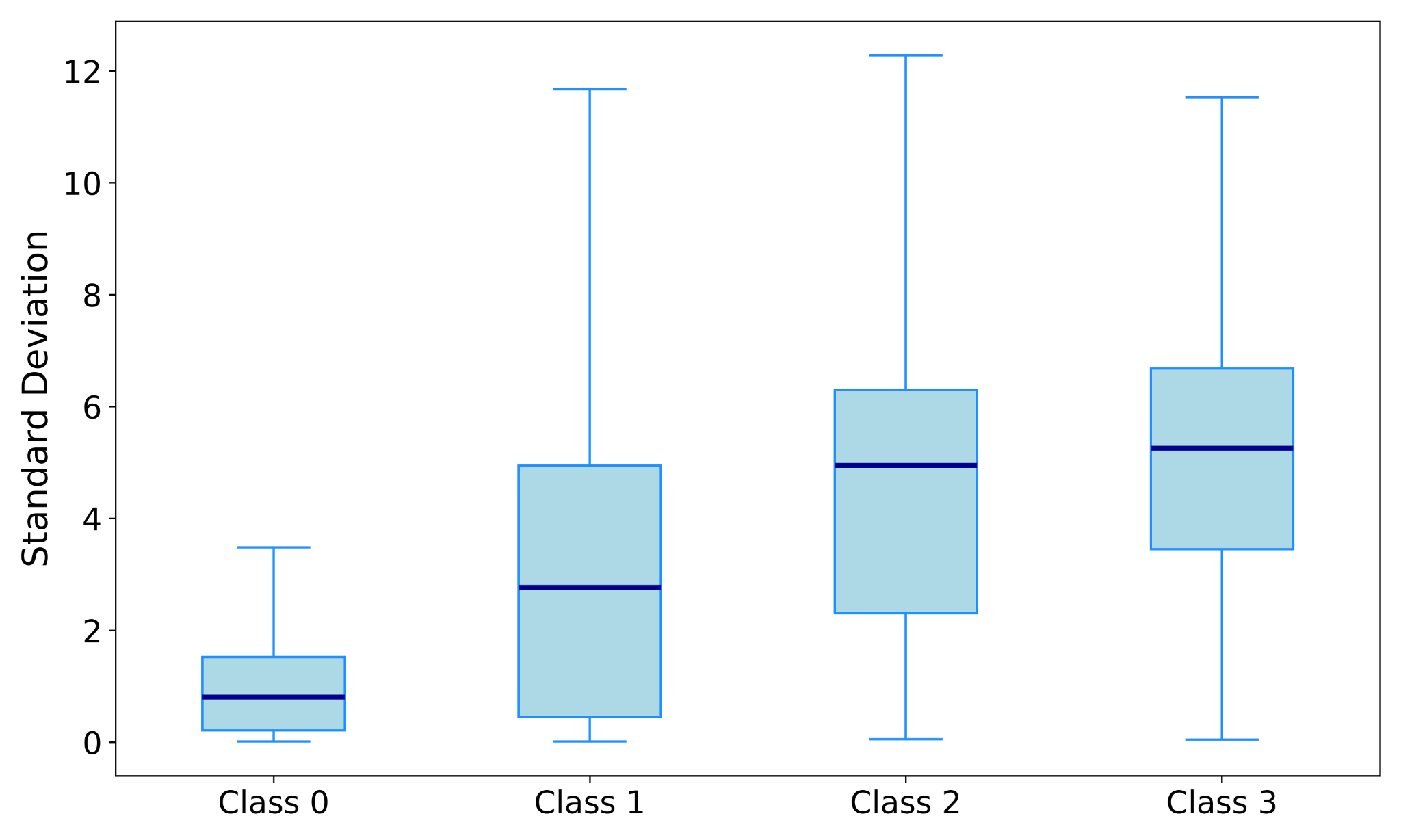}
\caption{Distribution of time-wise standard deviations by number of occupants. 
Box plots show median, quartiles, and whiskers (1.5$\times$IQR). Outliers are not 
displayed for visual clarity.}
\label{fig:boxplot_std}
\end{figure}

As shown in Figure~\ref{fig:boxplot_std}, the standard deviation increases with the number of occupants, clearly distinguishing human presence from the 0-person case. Based on this property, we designed two preprocessing methods that suppress static reflections and emphasize human-related fluctuations to improve spatial generalization.

\paragraph{Threshold-based Zeroing}
In the first method, the standard deviation across the 60-frame sequence is computed for each cell. If the standard deviation is below a predefined threshold $\tau$, the corresponding values are set to zero across all time steps. This effectively removes reflections from static objects or noise with low variance. For standard deviation map $\sigma(x, y)$ and threshold $\tau$, the transformation is formalized as:

\begin{equation}
X'_t(x, y) = \begin{cases}
0, & \text{if } \sigma(x, y) \leq \tau \\
X_t(x, y), & \text{otherwise}
\end{cases}
\end{equation}

for all $t \in \{1, 2, \dots, 60\}$.

The threshold value $\tau = 0.02$ was selected based on statistical analysis of the training data. For the no-person case (Label 0), the median standard deviation was 0.013, with 75\% of pixels showing standard deviations below 0.025. In contrast, when people were present (Label 1 and above), the standard deviation increased significantly, with median values of 0.046 or higher. The threshold of 0.02 was set higher than the typical values of the no-person class (median 0.013) and close to the 75th percentile (0.025), functioning as an optimal separation point that effectively suppresses static background while preserving dynamic changes caused by human presence.

\paragraph{Sigmoid-based Weighting}
The second method applies a continuous weighting scheme using a sigmoid function. After computing the standard deviation map $\sigma(x, y)$, a weight map $w(x, y)$ is generated as:

\begin{equation}
w(x, y) = \frac{1}{1 + \exp\left( -\frac{\sigma(x, y) - \tau}{s} \right)}
\end{equation}

where $\tau = 0.02$ serves as the midpoint of the sigmoid curve (same threshold as above), and $s = 0.01$ controls the steepness. The steepness parameter $s = 0.01$ was determined through systematic evaluation using multiple candidate values, with $s = 0.01$ achieving the highest Fisher Score and Adjusted Mutual Information (AMI) in the clustering analysis described in the next section. The weighted reflection amplitude is then computed as:

\begin{equation}
X'_t(x, y) = w(x, y) \cdot X_t(x, y) \quad \forall t \in [1, 60]
\end{equation}

Compared to binary zeroing, this approach allows for smoother suppression of low-saliency regions and better preserves marginal signals from partially moving targets. This is particularly beneficial for detecting subtle human movements near the decision boundary.

\subsubsection{Filtering Techniques for Temporal Noise Suppression}

In addition to amplitude-based preprocessing, we investigated frequency-domain filtering to mitigate systematic noise and drift that vary across environments. Human reflections show temporal fluctuations from respiration, heartbeat, and micro-movements, whereas static objects remain constant \cite{Li2013_TMTT_Healthcare,Chen2006_TAES_MicroDoppler,KimLing2009_TGRS}. These physiological signals typically occur at \SIrange{0.2}{0.5}{\hertz} (respiration), \SIrange{1}{2}{\hertz} (cardiac), and \SIrange{1}{3}{\hertz} (walking) \cite{Li2013_TMTT_Healthcare,Chen2006_TAES_MicroDoppler,PachiJi2005_StructEng}, though radar sensing often shifts them to lower observable frequencies due to propagation and system constraints \cite{Meta2007_FMCW_SignalProcessing,Skolnik2008_RadarHandbook}.

Given our sampling rate of \SI{8.57}{\hertz} (Nyquist limit \SI{4.29}{\hertz}), we evaluated two filters. The first is a fourth-order Butterworth band-pass (\SIrange{0.1}{0.5}{\hertz}) to capture low-frequency motion such as postural sway. The second is a two-stage high-pass scheme: an 8th-order filter removing drift below \SI{0.05}{\hertz}, followed by a 2nd-order filter for smoother attenuation in \SIrange{0.05}{0.1}{\hertz}. Their outputs are combined with optimized weights (0.7:0.3), selected for maximal Fisher Score and Adjusted Mutual Information (AMI) explained in Section~V.  

\subsubsection{AutoEncoder Architecture for Background Suppression}

\begin{figure}[htp]
\centering
\includegraphics[width=0.7\linewidth]{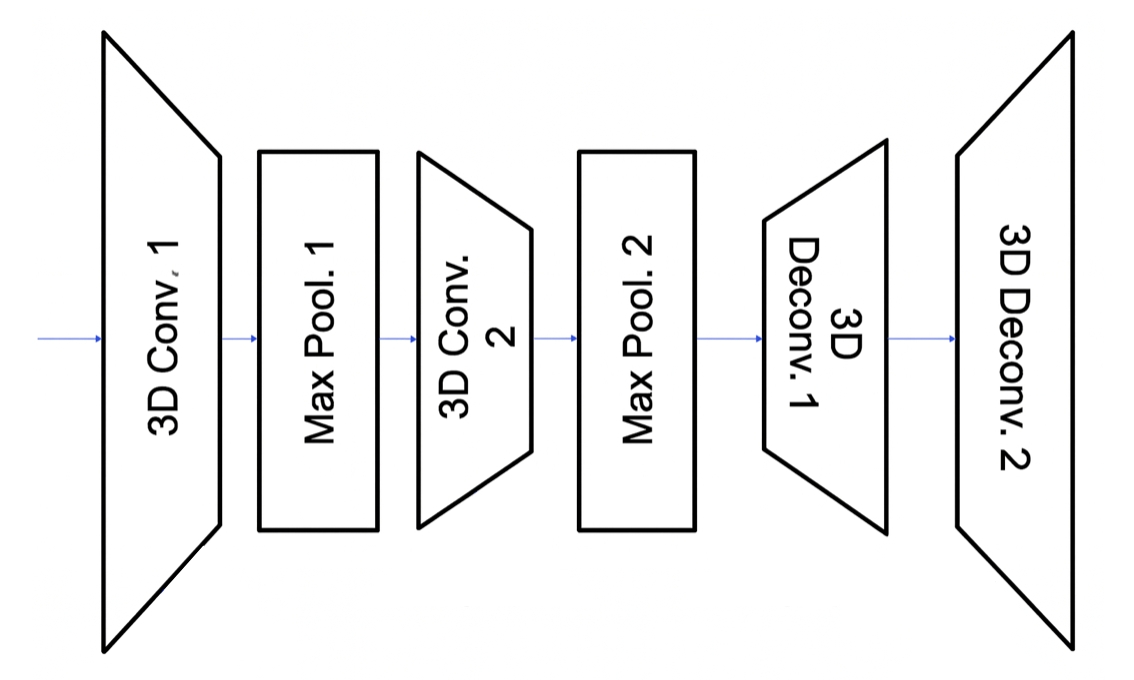}
\caption{Architecture of the 3D convolutional autoencoder.}
\label{fig:autoencoder_arch}
\end{figure}

To suppress static background reflections and extract human-induced motion, we designed a lightweight 3D convolutional autoencoder (Fig.~\ref{fig:autoencoder_arch}, Table~\ref{tab:autoencoder_config}). Compared to a CNN+LSTM approach, the 3D structure showed superior background suppression performance and was therefore adopted in this study.  

The model was trained in a self-supervised manner using 2,200 background-only samples (0-person class, Environment A) with MSE loss and the Adam optimizer ($10^{-3}$ learning rate). To avoid overfitting, the encoder and decoder each consist of only two stages. During inference, the reconstructed background from the autoencoder is subtracted from the original input, leaving residual components corresponding to human reflections \cite{Autoencoder_ICASSP2020,GraphAttnAE2023}.  

\begin{table}[htp]
\caption{Layer configuration and dimensional transformations \\
of the 3D convolutional autoencoder.}
\label{tab:autoencoder_config}
\centering
\footnotesize
\renewcommand{\arraystretch}{1.1}
\begin{tabular}{c p{2.4cm} c c}
\hline
\textbf{Stage} & \textbf{Layer Description} & \textbf{Input Shape} & \textbf{Output Shape} \\
\hline
Input & Input tensor & $(1,60,12,91)$ & $(1,60,12,91)$ \\
      & (normalized) &  &  \\
\hline
Encoder 1 & Conv3D & $(1,60,12,91)$ & $(8,60,12,91)$ \\
          & (1$\rightarrow$8, padding=1) + ReLU & & \\
          & MaxPool3D & $(8,60,12,91)$ & $(8,30,6,45)$ \\
\hline
Encoder 2 & Conv3D & $(8,30,6,45)$ & $(16,30,6,45)$ \\
          & (8$\rightarrow$16, padding=1) + ReLU & & \\
          & MaxPool3D & $(16,30,6,45)$ & $(16,15,3,22)$ \\
\hline
Decoder 1 & ConvTranspose3D & $(16,15,3,22)$ & $(8,30,6,44)$ \\
          & (16$\rightarrow$8, & & \\
          & stride=2) + ReLU & & \\
\hline
Decoder 2 & ConvTranspose3D & $(8,30,6,44)$ & $(1,60,12,91)$ \\
          & (8$\rightarrow$1, stride=$2^3$) & & \\
          & + Sigmoid & & \\
\hline
\end{tabular}
\end{table}

\subsection{Data Augmentation for Enhancing Spatial Generalization}

To improve the spatial generalization of the people-counting model across unseen environments, we implemented three types of data augmentation techniques specifically designed for radar signal characteristics. The effectiveness of each method was evaluated by training models on data from Environment A (with and without augmentation) and comparing their performance on test data from Environment B.

\paragraph{Symmetry-Based Flipping}
Each original radar clip consists of a 3D tensor of size $60 \times 12 \times 91$, representing 60 time frames of 2D range-azimuth maps. We applied deterministic flipping operations across spatial axes to create three augmented versions: azimuth (left-right) flipping, range (top-bottom) flipping, and both directions simultaneously. These augmentations were performed identically across all time frames to preserve temporal consistency~\cite{Iwana2021_TimeSeriesAug}, potentially increasing the dataset size by a factor of four.

\paragraph{Random Scaling}
To simulate variations in reflection intensities caused by environmental factors such as humidity, temperature, or subject-specific differences in radar cross-section~\cite{Sizov2013_TempHumidity,Fioranelli2015_RCS_Personnel}, we applied multiplicative scaling to each input sequence. During training, each tensor was scaled by a random factor sampled uniformly from the range [0.95, 1.05]. This range was chosen to introduce meaningful variations while avoiding distortion of the underlying signal characteristics. This process perturbs the overall amplitude distribution without distorting spatial-temporal structure, helping the model learn scale-invariant features.

\paragraph{Frame Dropping and Interpolation}
To emulate temporal inconsistencies such as frame loss or timing jitter that may occur in real-world radar systems ~\cite{Meta2007_FMCW_SignalProcessing}, we implemented a structured frame-dropping strategy. Three frames were randomly removed—one from each temporal third (frames 0–19, 20–39, 40–59). Each removed frame was replaced by the linear interpolation of its temporally adjacent frames: \( x_t = (x_{t-1} + x_{t+1}) / 2 \).
 This technique introduces local motion variations while maintaining global sequence integrity. Repeating this process with different dropout indices during training enabled fourfold augmentation.

\subsection{Transfer Learning for Environmental Adaptation}
To assess transfer learning for rapid deployment in new environments, models trained on Environment A were fine-tuned on Environment C, which contained 1,000 samples (250 per class). The dataset was split into 540 training, 60 validation, and 400 test samples.

We compared two settings: (1) direct testing of Environment A models on Environment C without adaptation, and (2) fine-tuning with Environment C data. Four training set sizes (100, 200, 400, 540 samples) were tested using stratified sampling to maintain class balance. Fine-tuning updated all parameters with a reduced learning rate of $1 \times 10^{-4}$, trained up to 100 epochs with early stopping (patience=10) \cite{Yosinski2014_TransferLearning}. All models were evaluated on the same 400-sample test set to ensure fair comparison. 

\section{Experimental Results}

\subsection{Evaluation Metrics}

All experiments were evaluated using the following metrics:

\paragraph{Clustering Metrics}
For clustering analysis, we employed Adjusted Mutual Information (AMI) and Fisher Score \cite{Vinh2010_AMI_JMLR,Duda2001_PatternClassification}. 

\textbf{Adjusted Mutual Information (AMI)} quantifies how closely the clustering result aligns with the ground-truth labels (e.g., number of people or environment). Unlike simple accuracy, AMI adjusts for chance agreement, providing a more reliable evaluation. It is calculated as:

\begin{equation}
\AMI(U,V) = \frac{\MI(U,V) - \mathbb{E}[\MI(U,V)]}
                 {\max(\ent(U), \ent(V)) - \mathbb{E}[\MI(U,V)]}
\end{equation}

where $\MI(U,V)$ denotes the mutual information between clusterings $U$ and $V$, 
and $\ent$ denotes entropy. The value ranges from 0 to 1, where a score near 0 indicates random-like behavior, and a score near 1 indicates perfect alignment.

\textbf{Fisher Score} measures how well features are separated between classes. It is defined as:
\begin{equation}
F_i = \frac{\sum_{j=1}^{c} n_j (\mu_{ji} - \mu_i)^2}{\sum_{j=1}^{c} n_j \sigma_{ji}^2}
\end{equation}
where $\mu_{ji}$ and $\sigma_{ji}^2$ are the mean and variance of the $i$-th feature in the $j$-th class, $\mu_i$ is the overall mean, and $n_j$ is the number of samples in class $j$. Higher values indicate stronger class separability.

\paragraph{Regression Metrics}
For regression performance, we report Root Mean Square Error (\textbf{RMSE}) and Mean Absolute Error (\textbf{MAE}), two standard metrics widely used to quantify prediction accuracy. RMSE penalizes larger errors more heavily, whereas MAE treats all errors equally. Lower values indicate better performance.

\subsection{Effectiveness of Preprocessing Techniques Based on Clustering Metrics}

We evaluated four preprocessing methods—threshold-based zeroing, sigmoid-based weighting, Butterworth filtering, and two-stage high-pass filtering—using Environment A data with person-count and layout labels. AMI and Fisher Score were computed separately for each labeling scheme. Table~\ref{tab:clustering_results} summarizes the results.

\begin{table}[htp]
\caption{Comparison of clustering performance across \\
different preprocessing methods.}
\label{tab:clustering_results}
\centering
\small
\renewcommand{\arraystretch}{1.2}
\begin{tabular}{p{1.7cm} c c c c c}
\hline
\textbf{Method} & \textbf{Metric} & \multicolumn{2}{c }{\textbf{Person Count}} & \multicolumn{2}{c}{\textbf{Layout Type}} \\
& & \textbf{Before} & \textbf{After} & \textbf{Before} & \textbf{After} \\
\hline
Threshold & AMI & 0.2805 & 0.1415 & 0.1497 & 0.0222 \\
Zeroing & Fisher & 0.1470 & 0.1248 & 0.1453 & 0.0116 \\
\hline
\textbf{Sigmoid} & \textbf{AMI} & \textbf{0.2805} & \textbf{0.3240} & \textbf{0.1497} & \textbf{0.0151} \\
\textbf{Weighting} & \textbf{Fisher} & \textbf{0.1470} & \textbf{0.3513} & \textbf{0.1453} & \textbf{0.0149} \\
\hline
Butterworth & AMI & 0.2139 & 0.0819 & 0.1550 & 0.0035 \\
Filter & Fisher & 0.1494 & 0.0008 & 0.1545 & 0.0007 \\
\hline
Two-stage & AMI & 0.2139 & 0.0686 & 0.1550 & 0.0038 \\
Filter & Fisher & 0.1494 & 0.0008 & 0.1545 & 0.0008 \\
\hline
\end{tabular}
\end{table}

Among the methods, sigmoid-based weighting was most effective, enhancing person-related separability while suppressing layout-dependent features. Threshold-based zeroing showed limited improvement, whereas both filtering methods excessively smoothed signals, degrading separability in both categories. Autoencoder-based preprocessing was excluded since its nonlinear transformation is not directly comparable with linear clustering metrics.

\subsection{End-to-End Performance Evaluation on Environment B}
To evaluate the effectiveness of preprocessing methods in real-world inference scenarios, we trained people-counting models on data from Environment A with six different preprocessing conditions, including the baseline model, and evaluated them on Environment B. Table~\ref{tab:preprocessing_improvement} shows the quantitative results, where sigmoid-based weighting achieved the best cross-environment performance with 50.1\% and 55.2\% improvements in RMSE and MAE respectively compared to the baseline model.

\begin{table}[htbp]
\centering
\caption{Preprocessing Methods Performance and Improvement Rates}
\label{tab:preprocessing_improvement}
\renewcommand{\arraystretch}{1.2}
\begin{tabular}{ p{3.5cm} c c c c }
\hline
\textbf{Method} & \multicolumn{2}{c }{\textbf{Environment A}} & \multicolumn{2}{c }{\textbf{Environment B}} \\

 & \textbf{RMSE} & \textbf{MAE} & \textbf{RMSE} & \textbf{MAE} \\
\hline
Baseline Model & 0.0845 & 0.0182 & 1.2474 & 0.8678 \\
\hline
Threshold Zeroing & 0.1462 & 0.0449 & 0.6801 & 0.4335 \\
\hline
\textbf{Sigmoid Weighting} & \textbf{0.1118} & \textbf{0.0303} & \textbf{0.6219} & \textbf{0.3888} \\
\hline
Butterworth Filter & 1.1179 & 0.9997 & 1.1180 & 0.9998 \\
\hline
Two-stage Filter & 1.1180 & 1.0000 & 1.1180 & 1.0000 \\
\hline
Autoencoder & 0.1597 & 0.0503 & 0.7617 & 0.5455 \\
\hline
\end{tabular}
\end{table}

\subsection{Performance Evaluation of Data Augmentation Techniques}

To evaluate cross-environment generalization, models trained on Environment A with different augmentation strategies were tested on Environment B (Table~\ref{tab:data_augmentation}). Symmetry-based flipping yielded the best RMSE improvement, while random scaling was most effective for MAE. Frame dropping with interpolation showed only minor gains. Overall, the improvements from data augmentation were modest (less than 10\%), especially when compared with the substantial gains achieved by sigmoid weighting in preprocessing.

\begin{table}[htbp]
\centering
\caption{Performance Comparison of Data Augmentation Techniques}
\label{tab:data_augmentation}
\renewcommand{\arraystretch}{1.2}
\begin{tabular}{ p{3.5cm} c c c c }
\hline
\textbf{Method} & \multicolumn{2}{c }{\textbf{Environment A}} & \multicolumn{2}{c }{\textbf{Environment B}} \\

 & \textbf{RMSE} & \textbf{MAE} & \textbf{RMSE} & \textbf{MAE} \\
\hline
Base Model & 0.0845 & 0.0182 & 1.2474 & 0.8678 \\
\hline
Symmetry-Based Flipping & 0.0858 & 0.0221 & 1.1887 & 0.8349 \\
\hline
Random Scaling & 0.0621 & 0.0130 & 1.1973 & 0.7917 \\
\hline
Frame Dropping and Interpolation & 0.0623 & 0.0102 & 1.2133 & 0.8285 \\
\hline
\end{tabular}
\end{table}

\subsection{Effectiveness of Transfer Learning}

Using sigmoid weighting as the baseline, we tested transfer learning from Environment A to Environment C with varying amounts of target data (Table~\ref{tab:transfer_learning_performance}). 

Results show that transfer learning greatly mitigates domain shift: even with only 100 samples, performance improved by more than half, and with 540 samples, errors were reduced by over 80–90\%. The improvements scaled with data size, demonstrating that transfer learning enables substantial adaptation to new environments with limited data.

\begin{table}[htbp]
\centering
\caption{Transfer Learning Performance with Different Training Data Sizes}
\label{tab:transfer_learning_performance}
\renewcommand{\arraystretch}{1.2}
\begin{tabular}{ p{3.5cm} c c c c }
\hline
\textbf{Method} & \multicolumn{2}{c }{\textbf{Environment A}} & \multicolumn{2}{c }{\textbf{Environment C}} \\
 & \textbf{RMSE} & \textbf{MAE} & \textbf{RMSE} & \textbf{MAE} \\
\hline
No Transfer & 0.1118 & 0.0303 & 0.6963 & 0.4422 \\
\hline
100-Dataset Transfer Learning & - & - & 0.3107 & 0.1655 \\
\hline
200-Dataset Transfer Learning & - & - & 0.2148 & 0.1038 \\
\hline
400-Dataset Transfer Learning & - & - & 0.1706 & 0.0643 \\
\hline
540-Dataset Transfer Learning & - & - & 0.1245 & 0.0386 \\
\hline
\end{tabular}
\end{table}

\section{Discussion}

\subsection{Analysis of Preprocessing Techniques}

Our comprehensive evaluation of various preprocessing approaches for improving spatial generalization performance in radar-based people counting revealed significant performance differences between methods.

\subsubsection{Filtering-Based Methods: Limitations and Challenges}

Frequency-domain filtering approaches (Butterworth band-pass filter and two-stage high-pass filter) demonstrated substantial limitations in both clustering metric evaluation and end-to-end performance assessment. AMI and Fisher Score analysis revealed that these methods excessively removed not only environment-specific information but also discriminative features related to human presence. In end-to-end evaluation, both filtering methods showed RMSE values exceeding 1.11 compared to the baseline model (range 0.08-1.25), resulting in lower prediction accuracy than the baseline in both Environments A and B.

This poor performance can be attributed to the complex nature of radar reflections in indoor environments. Human presence generates multi-path reflections involving interactions with surrounding objects such as chairs, desks, and whiteboards, resulting in signal components distributed across various frequency bands rather than concentrated in specific ranges~\cite{Yarovoy2006_UWB_Indoor}. Consequently, frequency-based filtering approaches face fundamental limitations in distinguishing between human-related and background reflections.

Furthermore, our radar dataset contained temporal irregularities including slight acquisition interval variations and duplicate timestamps. Such timing misalignments likely degraded the precision of frequency-domain filters, as accurate temporal sampling is crucial for reliable frequency analysis. Similar timing alignment issues have been reported in other radar sensing studies~\cite{Hazra2018_Robust_Gesture}, suggesting that filtering-based preprocessing may have inherent robustness limitations in practical deployment scenarios where perfect timing control is difficult to achieve.

\subsubsection{Statistical Amplitude-Based Methods: Superior Performance}

In contrast, the sigmoid-based amplitude weighting method demonstrated the most consistent improvement in distinguishing between different numbers of people while suppressing environmental variability. This approach achieved 50.1\% improvement in RMSE and 55.2\% improvement in MAE on Environment B compared to the baseline model. The threshold-based zeroing method also showed meaningful improvements, though with more modest effects (45.5\% RMSE improvement, 50.0\% MAE improvement).

The superior performance of sigmoid weighting over threshold zeroing can be attributed to its continuous weighting scheme, which prevents the complete elimination of human-related signal components that might occur near decision boundaries in binary thresholding. This preservation of boundary signals is particularly important for detecting subtle human movements or partially occluded targets.

\subsubsection{Deep Learning-Based Preprocessing: Overfitting Concerns}

The autoencoder-based approach did not outperform statistical methods despite its theoretical capability to learn complex spatial-temporal patterns. The autoencoder achieved only moderate improvements (39.0\% RMSE improvement, 37.1\% MAE improvement on Environment B), significantly lower than sigmoid weighting. This limitation likely stems from the model's susceptibility to overfitting to Environment A's specific characteristics, limiting its effectiveness in Environment B with different spatial layouts ~\cite{Ghifary2016_Autoencoder_Domain}. 

\subsection{Data Augmentation: Limited but Measurable Impact}

Among data augmentation techniques, symmetric flipping showed the most favorable improvement in RMSE (4.7\% improvement), while random scaling was most effective for MAE reduction (8.8\% improvement). Frame dropping and interpolation demonstrated more limited effectiveness with 2.7\% RMSE and 4.5\% MAE improvements.

However, all data augmentation methods showed substantially smaller improvement rates compared to sigmoid weighting preprocessing. This limited effectiveness suggests that data augmentation alone has inherent constraints in generating diverse spatial patterns sufficient to bridge significant environmental gaps. Sigmoid weighting preprocessing likely proved more effective because it directly addresses the fundamental challenge of separating human-related signals from environmental noise, rather than simply increasing data variety.

\subsection{Transfer Learning for Significant Domain Shifts}

For Environment~C, which exhibits substantially different spatial structures, transfer learning was essential to maintain acceptable performance. Even with a small dataset of only 100 samples, transfer learning achieved improvements of 55.4\% in RMSE and 62.5\% in MAE compared to direct deployment. With 540 samples, the improvements further increased to 82.1\% and 91.3\%, respectively.

These substantial gains indicate that while sigmoid-based weighting is effective for moderate environmental variations, transfer learning becomes indispensable when spatial characteristics differ significantly~\cite{CrossEnvHAR2022_JSEN}. Moreover, our lightweight deep learning model can be fully retrained with a small amount of target-domain data, achieving strong fine-tuning effectiveness and thereby demonstrating particularly high practical utility for transfer learning.

\subsection{Practical Implications and Future Directions}

Based on these findings, we conclude that when constructing deep learning models for radar sensing using amplitude fluctuations as features, the optimal approach would be to implement sigmoid weighting preprocessing to provide robustness against moderate environmental changes, and employ transfer learning for deployment in significantly different environments.

\section{Conclusion}

This study systematically evaluated methods to enhance spatial generalization in deep learning-based people counting with FMCW MIMO radar. Sigmoid-based amplitude weighting preprocessing proved most effective for moderate environmental variations, while transfer learning was indispensable for larger spatial changes, achieving strong adaptation with minimal data. These findings, from the first comprehensive evaluation of such methods, provide practical guidelines for building radar sensing systems that sustain accuracy across diverse environments and enable real-world deployment.

\section*{Acknowledgment}
The authors gratefully acknowledge Prof. Tei Sigaku, Emeritus Professor at the University of Aizu, and Dr. Aisaku Nakamura for their insightful guidance and discussions. The authors also appreciate SoftBank Corp. for offering the research environment and technical support.

\bibliographystyle{IEEEtran}
\bibliography{refs}

\end{document}